\documentclass[10pt,twocolumn,letterpaper]{article}

\usepackage{wacv}
\usepackage{times}
\usepackage{epsfig}
\usepackage{graphicx}
\usepackage{amsmath}
\usepackage{amssymb}
\usepackage{booktabs}
\usepackage{comment}
\usepackage{subcaption}
\usepackage{enumitem}
\usepackage{eso-pic,graphicx}
\usepackage[accsupp]{axessibility}
\DeclareMathOperator*{\argmin}{arg\,min}

\def\wacvPaperID{447} 

\wacvalgorithmstrack   

\wacvfinalcopy 

\ifwacvfinal
\usepackage[breaklinks=true,bookmarks=false]{hyperref}
\else
\usepackage[pagebackref=true,breaklinks=true,colorlinks,bookmarks=false]{hyperref}
\fi

\begin{document}

\title{Continual Learning with Dependency Preserving Hypernetworks}

\author{Dupati Srikar Chandra$^{1}$  Sakshi Varshney$^{1}$  P.K. Srijith$^{1}$  
 Sunil Gupta$^{2}$\\
{}$^1$Indian Institute of Technology, India {}$^2$Deakin  University, Australia\\
{\tt\small \{ai20resch11004,cs16resch01002\}@iith.ac.in srijith@cse.iith.ac.in sunil.gupta@deakin.edu.au }\\
}
\maketitle
\thispagestyle{empty}

\begin{abstract}
   Humans learn continually throughout their lifespan by accumulating diverse knowledge and fine-tuning it for future tasks. When presented with a similar goal, neural networks suffer from catastrophic forgetting if data distributions across sequential tasks are not stationary over the course of learning. An effective approach to address such continual learning (CL) problems is to use hypernetworks which generate task dependent weights for a target network. However, the continual learning performance of existing hypernetwork based approaches are affected by the assumption of independence of the weights across the layers in order to maintain parameter efficiency. To address this limitation, we propose a novel approach that uses a dependency preserving hypernetwork to generate weights for the target network while also maintaining the parameter efficiency. We propose to use recurrent neural network (RNN) based hypernetwork that can generate layer weights efficiently while allowing for dependencies across them. In addition, we propose novel regularisation and network growth techniques for the RNN based hypernetwork to further improve the continual learning performance. To demonstrate the effectiveness of the proposed methods, we conducted experiments on several image classification continual learning tasks and settings. We found that the proposed methods based on the RNN hypernetworks outperformed the baselines in all these CL settings and tasks.
\end{abstract}


\section{Introduction}
There are various applications where a computational system has to learn from a stream of data continually and adapt to the environment by using knowledge gained from past experiences. For example, Autonomous agents in the real world have to learn over continuous streams of data and need to remember the information from various non-stationary distributions without forgetting~\cite{1}. Deep neural networks have achieved high performance on many images classification benchmarks, comparable or even better than humans. However, while learning a new task, these networks forget the knowledge gained from previous ones. 
The process of forgetting knowledge or information gained from previous tasks due to drastic weight alteration while learning new tasks is known as \textit{catastrophic forgetting}~\cite{38}. If data distributions across the sequential tasks is not stationary, the weights of the model alter drastically to classify the new task, leading to forgetting of knowledge gained from previous tasks. Ideally, performance of a newly learnt task should not have an impact on previous one (or vice versa). To overcome forgetting, computational systems or agents, on one hand, should be plastic to acquire new information and refine old information based on continuous input and, on the other hand, it should be stable to prevent the novel input from interfering with old information. This is referred to as \textit{plasticity-stability dilemma}~\cite{6,7}. \textit{Continual learning} aims to develop machine learning and deep learning models which are stable enough  to retain information learnt from old tasks but also has the required plasticity to learn new tasks~\cite{1}.

Continual learning techniques for neural networks has gained significant attention recently~\cite{1}. Several continual learning methods have been proposed to avoid forgetting in neural networks. These can be broadly categorised into three approaches, \textit{Regularisation techniques}~\cite{13,32},  \textit{dynamic architecture methods}~\cite{24,31} and \textit{replay based methods} ~\cite{18,22}. 
Recently, \textit{Continual learning with hypernetworks} has shown very promising results in dealing with forgetting at a meta level by generating task specific weights~\cite{14,4,5}. Hypernetwork is a meta neural network which generates parameters for the main network associated with the task (for instance, classification or regression network) by considering some task related information. During training, instead of directly trying to update parameters of the main network, hypernetwork parameters which generates them are updated. But generating the entire main network parameters will require a larger hypernetwork as it requires generating a very high dimensional output. Instead, chunked hypernetworks~\cite{4,5} generate them in  smaller chunks (chunk referred to subset of main network weights) multiple times iteratively using smaller hypernetwork that is reusable and also help in model compression significantly.
A key limitation of chunked hypernetworks is that it assumes the weight matrices associated with the chunks to be independent, which affects the continual learning performance significantly. Chunked hypernetwork~\cite{5} uses a feed forward neural network to generate weights without considering dependencies across chunks. 
We propose to use \textit{recurrent neural networks (RNNs)}, in order to capture  dependencies in weight matrix generation across the chunks. But standard \textit{RNNs} suffer from vanishing gradient problems and may not be able to remember dependencies for a long time. So, a variant of \textit{RNN}, \textit{LSTM}~\cite{13} has been used to remember the weight matrix dependencies over a  longer duration.
Thus, we propose \textit{LSTM based hypernetwork} that can efficiently generate weights of the main network while also maintaining the dependencies across the chunks. 

While learning multiple tasks in sequential manner, the hypernetwork should remember the knowledge gained from previous tasks and should also be able to forward that knowledge to the upcoming tasks. To achieve this, we propose a novel hypernetwork regularisation technique, \textit{Importance Weighted Regularisation (\textit{IWR})} that can further improve the performance of \textit{hypernetwork based continual learning}  by enabling forward transfer while also retaining previous information. \textit{IWR} considers the  importance of the main network parameters and allows more flexibility to the hypernetwork to adapt to the new task by considering this importance. We also propose a \textit{network growth } technique for the  \textit{LSTM} based hypernetworks for continual learning. The approach is based on the idea that the dependence among the main network parameters across tasks remains the same while their exact values could be different. This is achieved by sharing the weights associated with the hidden states in the \textit{LSTM}  across the tasks. We still learn the input specific weights for each task separately.  This approach improves the performance of continual learning with no extra regularisation and accelerates the model training. Our experimental results on real world data show that the proposed approaches along with \textit{LSTM} based hypernetwork can effectively mitigate  catastrophic forgetting  and  significantly improve  the continual learning  performance.

Our main contributions can be summarised as follows.
\begin{itemize} [noitemsep]
\item We propose a novel dependency preserving \textit{LSTM} hypernetwork for continual learning.
\item We propose a novel regularisation technique for hypernetwork based continual learning and a network growth technique specifically for the \textit{LSTM} based hypernetwork which does not require regularisation.
\item We demonstrate the improvement in continual learning performance of the proposed approaches through experiments on  several image classification data sets and for different CL settings.
\end{itemize}

\section{Related Work}
Several approaches were proposed recently for continual learning and to deal with catastrophic forgetting. Conceptually, these approaches are classified based on replaying the stored examples, methods that expand the model on seeing the new task and methods that regularise the parameter shift by retaining the network~\cite{1,2}. 
\textit{Regularisation based approaches}~\cite{21,10,20,11} avoid forgetting by imposing constraints on the update of parameters. With the advantage of no extra memory requirement, regularisation approaches are used for a broader variety of applications that have constraints on memory, computational resources, and data privacy. \textit{Elastic weight consolidation(EWC)}~\cite{10} and \textit {Synaptic intelligence(SI)}~\cite{11} are most well-known approaches, proposed to mitigate forgetting by constraining the update of important parameters of the task. It imposes a quadratic penalty on the difference between old and new parameters which helps to slow down the learning of new tasks by updating old parameters.  But experiments in~\cite{34} showed that \textit{EWC} is not effective in learning new classes incrementally.

\textit{Replay based methods} \cite{24,23,22} alleviate catastrophic forgetting by replaying old examples while learning new tasks. These methods either store examples from previous tasks or, generate synthetic examples from trained generative models from learnt feature space. \textit{Variational autoencoders (VAEs)} and \textit{Generative adversarial networks (GANs)} are used to generate samples from feature space. \textit{iCaRL}~\cite{22}, stores a subset of samples per class in fixed memory and selected examples should best approximate the class means in feature space. But as the number of tasks grows, samples per class stored will become too few to ensure required performance. Various other approaches~\cite{27,25,26} have been proposed which generate samples for old tasks instead of storing them and these are replayed while learning a new task.

\textit{Dynamic architecture based methods} provide a solution towards continual learning by growing or updating its model structure for each task~\cite{29,28}. \textit{Progressive neural network (PNNs)}~\cite{30} grow their architecture by expanding the network statically with new modules. Forgetting can be circumvented by adding lateral connections from previous modules. Instead of growing the network structure statically, \textit{Dynamically expandable network (DEN)} grows network architecture for each task with only a limited number of units and identifies neurons that are important for the new task  and  train them selectively~\cite{28}.

Recently, hypernetwork based approaches have been proposed~\cite{4,34,12,5} which has the advantage of having a constrained search space compared to the  main network. Hypernetwork based techniques use a secondary neural network to generate the parameters of the main network and deal with forgetting at the hypernetwork level.  To keep the size of the hypernetwork to be small compared to the main network, they generate the weights of the main network in small fixed sized chunks~\cite{4}.   However, we notice that the process of independent generation of chunk weights ignores the dependence among main network parameters and thus affects the continual learning performance. To overcome this, we propose a \textit{LSTM} based hypernetwork that can generate weights in smaller chunks while also maintaining dependency across them. Due to their ability to capture dependencies, \textit{LSTMs} were used for learning from sequential data.  Recently, continual learning methods for recurrent neural networks based on existing regularisation technique and hypernetworks were proposed in~\cite{14}.  
In contrast to the work in~\cite{14}, where the aim is to model continual learning for tasks involving sequence data  using \textit{RNNs}  with existing CL techniques, the goal of this paper is to develop novel  continual learning methodology based on \textit{RNNs} (specifically \textit{LSTMs})  by treating them as a hypernetwork. In addition, we also introduce novel continual learning techniques for such \textit{LSTM}  based hypernetworks such as importance weighted regularisation and network growth.

\section{LSTM Hypernetwork and Regularization Techniques for Continual Learning}

In many realistic real world learning scenarios, the tasks arrive in sequential manner. Continual learning aims to learn from a sequence of tasks where the data of all the tasks are not available at once and we have a fixed memory size.  We assume that we are given a sequence of $K$ tasks,  where  each task \( t \in \mathcal{T} = \{1,\ldots, K\} \) contains input \(X^{t}\)=\(\{x_{j}^{t}\}_{j=1}^{n_t}\) and target label  \(Y^{t}\)=\(\{y_{j}^{t}\}_{j=1}^{n_t}\) where \(n_t\) being the number of  samples in task t. The goal of main network $m$ is to learn a function $f_m^t{(\cdot,{\Theta_{m}^t}}):X^{t}\to{Y^{t}}$ with parameters \(\Theta_m^t\) associated with task $t$.  While learning a task $t$ we have access to only observations of current task $t$ but no access to data of previous tasks.

We can learn \(\Theta_m^t\) separately for each task, but it results in a linear growth in number of parameters and the fixed sized memory will not be sufficient to store them. If we maintain the main network parameter to be same across all the tasks, the parameter values will get overwritten by the new task data which will results in catastrophic forgetting. To learn continuously over the tasks without requiring a linear growth in parameters, hypernetworks are proposed to generate the main network parameters for each task.  The hypernetwork $h$ learns a function $f_h{(\cdot,\Theta_h}): \mathbf{e^t} \to {\Theta_m^t}$ to generate the task specific parameter \( \Theta_m^t \) given a task embedding $\mathbf{e^t}$ using   trainable parameters \( \Theta_h \). 
  
Generating the high dimensional main network parameters all at once requires a very large hypernetwork with a large number of outputs and is computationally costly to train them. 
With the motive of reducing the number of trainable parameters in hypernetworks, chunked hypernetworks~\cite{4,5} are proposed to generate weights in smaller chunks (subsets of the weight matrix of main network)  by  reusing the same hypernetwork $f_h$ multiple times with different chunk embeddings.
 
Thus, hypernetwork \(f_h\) with parameters \(\Theta_h\) takes task embedding \(\mathbf{e^t}\) and chunk embeddings $\mathbf{c}$ = $\{\mathbf{c}_1, \ldots, \mathbf{c}_{n_c}\}$   as input to generate set of main network weights,
$\Theta_{m}^t$ = $f_h(\mathbf{e^t}, \mathbf{c}, \Theta_h)$ = $\{{f_h}(\mathbf{e^t},{\mathbf{c}}_1, \Theta_h),{f_h}(\mathbf{e^t},{\mathbf{c}}_2, \Theta_h), \ldots,{f_h}(\mathbf{e^t},{\mathbf{c}}_{n_c}, \Theta_h)\}$,
where $n_c$ being the number of chunks.
hypernetworks can still suffer from catastrophic forgetting when hypernetwork parameters are updated to generate main network parameters for the new
task. In order to overcome this forgetting, an additional regularisation term is used while learning the parameters of the hypernetwork for a new task along with task specific loss~\cite{5}.


Chunked hypernetworks generate weights of the main network in smaller chunks using a hypernetwork considering the task embedding and chunk embedding. We observe that they do not consider sequential nature and inter-dependence of weights between the chunks. We note that the chunked hypernetworks make conditional independence assumptions on the weights of the main network. Consequently, if we consider a probability distribution over weights, it gets decomposed over chunks as 
\begin{small}
$P({\Theta_{m}^{t,1}},{\Theta_{m}^{t,2}},{\ldots},{\Theta_{m}^{t,n_c}}|{\mathbf{e^t},\textbf{c}})  =  P({\Theta_{m}^{t,1}}|{\mathbf{e^t},\mathbf{c}_1}) \times P({\Theta_{m}^{t,2}}|{\mathbf{e^t},\mathbf{c}_2})\ldots P({\Theta_{m}^{t,n_c}}|{\mathbf{e^t},\mathbf{c}_{n_c}})$.
 \end{small}
Assumption of independence across the chunks does not usually hold and can affect the main network parameter generation and  the performance of the continual learning.

  \subsection{LSTM Hypernetworks}

 To capture the interdependence in the chunk weights and to be parameter efficient at the same time, we propose a recurrent neural network (\textit{RNN})  and in particular long short term memory (\textit{LSTM})~\cite{13} based hypernetwork called  \textit{LSTM\_NET}.  \textit{LSTM\_NET} is capable of  generating weights for main network in smaller chunks while also maintaining the dependencies across the chunks. \textit{LSTMs} are sequence models capable of capturing long range dependencies and will be able to  generate a chunk weight depending on the weights associated with  the preceding chunks. Consequently, it models the  joint probability over the main network parameters as

\begin{small}
\begin{equation}
\begin{split}
\label{eq:3}
 \hspace{-0.1in} P({\Theta_{m}^{t,1}},{\Theta_{m}^{t,2}},{\ldots},{\Theta_{m}^{t,n_c}}|{\mathbf{e^t},\textbf{c}}) & = P({\Theta_{m}^{t,1}}|{\mathbf{e^t},\mathbf{c}_{1}}) \times  \\&   \mkern-36mu \mkern-36mu \mkern-36mu \mkern-36mu \mkern-36mu \mkern-36mu P({\Theta_{m}^{t,2}}|{\Theta_{m}^{t,1}},{\mathbf{e^t},\mathbf{c}_{2}}) \ldots    P({\Theta_{m}^{t,n_c}}|{\Theta_{m}^{t,1}}\ldots  {\Theta_{m}^{t,{n_c-1}}}, {\mathbf{e^t},\mathbf{c}_{n_c}})
\end{split}
\end{equation}
\end{small}

\begin{figure}[t]
\hspace{0.8 in}\centerline{\includegraphics[height=65mm,width=130mm]{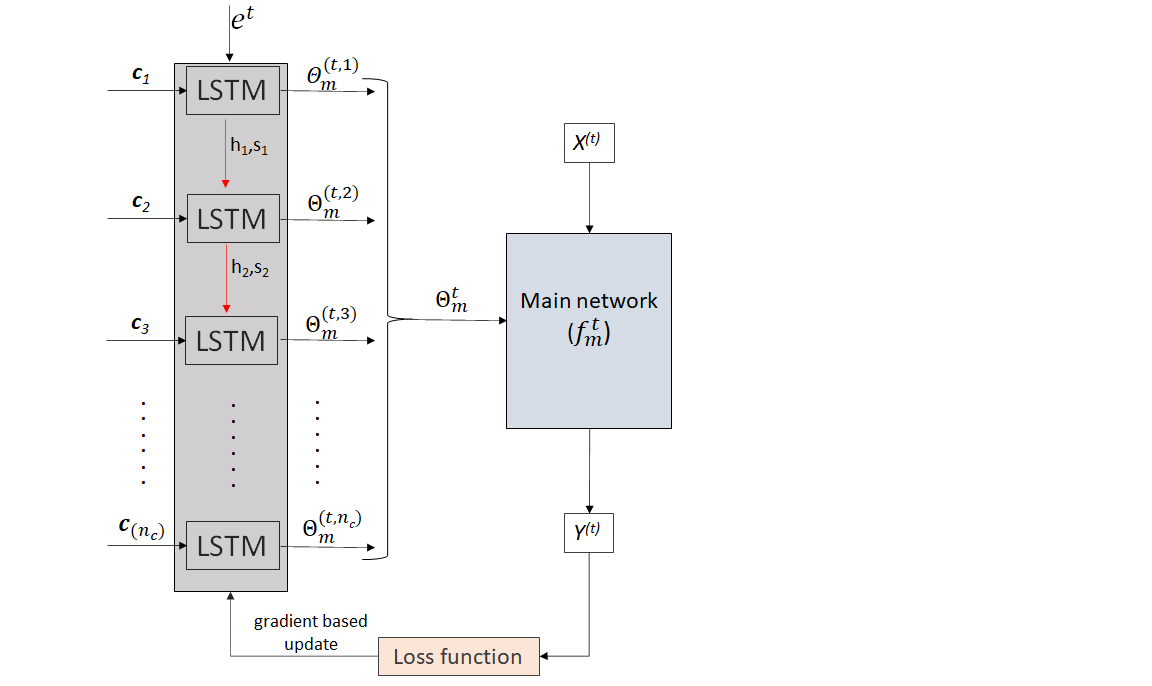}}
\caption{\textbf{LSTM based Hypernetwork }  can generate main network  weights \(\Theta_{m}^t\)  for task $t$ in smaller chunks by using same hypernetwork iteratively but also by  maintaining the dependencies across them using \textit{LSTM}.  }
\label{fig0}
\end{figure}
The proposed \textit{LSTM} hypernetwork uses  hidden state $\mathbf{h_{j-1}^{t}}$ and cell state $\mathbf{s_{j-1}^{t}}$ of preceding chunk along with  current chunk embedding to generate the chunk weights.  The following \textit{LSTM} operations are used to generate the chunk weights assuming a single layer network where it takes task embedding $\mathbf{e^t}$ and  chunk embedding $\mathbf{c_j}$ as input. 
\begin{align*}
&\mathbf{i_j^{t}} =\sigma(\mathbf{w_i}\times{\langle\mathbf{e^t},\mathbf{c_j}\rangle}+\mathbf{u_i}\times \mathbf{h_{j-1}^{t}} ) \\
&\mathbf{f_j^{t}} =\sigma(\mathbf{w_f}\times{\langle\mathbf{e^t},\mathbf{c_j}\rangle}+\mathbf{u_f}\times\mathbf{h_{j-1}^{t} }) \\
&\mathbf{o_j^{t}} =\sigma(\mathbf{w_o}\times{\langle\mathbf{e^t},\mathbf{c_j}\rangle}+\mathbf{u_o}\times\mathbf{h_{j-1}^{t}} ) \\
&\mathbf{g_j^{t}} = Tanh(\mathbf{w_g}\times{\langle\mathbf{e^t},\mathbf{c_j}\rangle}+\mathbf{u_g}\times\mathbf{h_{j-1}^{t}} )\\
&\mathbf{s_j^{t}} = \mathbf{f_j^{t}} \odot \mathbf{s_{j-1}^{t}}  + \mathbf{i_j^{t}}  \odot \mathbf{g_j^{t}} \\
&\mathbf{h_j^{t}}  = \mathbf{o_j^{t}} \odot \tanh(\mathbf{s_j^t})\\
&\Theta_m^{(t,j)}=\mathbf{h_j^{t}} W 
\end{align*}

where \(\mathbf{f,i,o,g}\) are forget gate, input gate, output gate, and cell gate respectively. \(\sigma\) is the sigmoid function, and \(\odot\) is the Hadamard product. \textbf{w, u} are the weights associated with the input and  hidden states respectively with subscript denoting the corresponding gate.  \(\Theta_m^{(t,j)}\) is the $j^{th}$ chunk of main network weights generated by \textit{LSTM} hypernetwork. Here, \(W\in R^{d_1\times d_2}\) are the weights of the feed forward layer producing the chunk weights. 

We learn  the \textit{LSTM} parameters $\Theta_h$ when presented with data from task $T$ by minimising the following loss consisting of task specific loss and regularization loss~\cite{5}.

\begin{small}
\begin{equation}
\begin{split}
\label{eq:1}
\hspace{-0.15in}
\argmin_{\Theta_h}  \mathcal{L}_{total}  & =   \mathcal{L}_{task}(\Theta_{h},\mathbf{e}^{T}, \mathbf{c}, X^{T},Y^{T}) 
\\ 
 & \mkern-36mu \mkern-36mu \mkern-18mu \mkern-18mu 
 +\frac {\beta} { T-1} \sum_{t=1}^{T-1}{||}f_{h}(\mathbf{e}^{t},\mathbf{c}, \Theta_{h}^ {*})-f_{h}(\mathbf{e}^{t},\mathbf{c},\Theta_{h}+\Delta\Theta_{h}){||^2}
 \end{split}
\end{equation}
\end{small}
where $f_h(\mathbf{e^t}, \mathbf{c}, \Theta_h) = \Theta_{m}^t$ denotes all the parameters of the main network for task $t$, \(\Theta_{h}^{*}\) are parameters of the hypernetwork before learning task $T$, \(\beta\) is a regularisation constant which balances task specific loss and regularization loss, and \(\Delta\Theta_h\) is change in direction of the weights of the hypernetwork evaluated on the task-specific loss. 
The task-specific loss $\mathcal{L}_{task}$ is the loss associated with the task (for e.g., cross-entropy loss for classification). The regularization loss constraints the hypernetwork learnt on the new task to generate main network parameters similar to the ones generated by previously learnt hypernetwork. Hypernetwork parameters \(\Theta_{h}\) and chunk embeddings are learnt by minimising the total loss $\mathcal{L}_{total}$, and the task embeddings are learnt using $\mathcal{L}_{task}$ alone using backpropagation.

\subsection{Importance Weighted Regularisation}

We propose a novel regularisation technique for continual learning in hypernetworks which provides more flexibility to the hypernetwork to adapt to the new task compared to the regularisation in \eqref{eq:1}. The proposed \textit{importance weighted regularisation (IWR)} updates the hypernetwork parameters based on the importance of the  parameters associated with the main network for each task. \textit{IWR} requires  the hypernetwork to generate only important main network parameters associated with old tasks and not all the main network parameters. We achieve this by considering the fisher information score of the main network parameters in the regularisation term in \eqref{eq:1}.  This will enforce hypernetwork to give importance to significant main network parameters during generation of main network weights. Meanwhile, it provides flexibility to the hypernetwork to adapt its parameters more freely to the new task as it is not constrained to generate all the main network parameters with equal importance. The objective function considering the \textit{IWR} regularisation  is defined as 

\begin{small}
\begin{equation}\label{eq:4}
\begin{split}
\argmin_{\Theta_h} \mathcal{L}_{total} & =  \mathcal{L}_{task}(\Theta_{h},\mathbf{e}^{T},\mathbf{c},X^{T},Y^{T}) \\ 
 & \mkern-36mu \mkern-36mu \mkern-36mu \mkern-6mu +{ \frac {\beta} {T-1} \sum_{t=1}^{T-1}\sum_{i}
 FI^{t}_{i}({{f_h}_i}(\mathbf{e}^{t},\mathbf{c},\Theta_{h}^ {*})-{f_h}_i(\mathbf{e}^{t},\mathbf{c},\Theta_{h}+\Delta\Theta_{h}))^2 } 
\end{split}
\end{equation}
\end{small}
where the first term
\(\mathcal{L}_{task}\) is task specific loss and the second term is the \textit{IWR} term that  regularises the hypernetwork parameters to avoid forgetting.  The \textit{IWR} term  uses \({FI}^{t}\) which  is the fisher information matrix (defined below) over the main network parameters associated with task $t$.  \({FI}^{t}\) provides the importance of the main network parameters for task $t$, and the index $i$ iterates over all the main network parameters. ${f_h}_i(\mathbf{e}^{t},\mathbf{c},\Theta_{h})$ denotes the $i^{th}$ main network parameter generated by the hypernetwork. We can observe that if the Fisher information associated with the $i^{th}$ main network parameter is high (implying that this parameter is important), then hypernetwork is required to generate it exactly while it does not have to do the same for the unimportant parameters. Thus, \textit{IWR} provides more flexibility to the hypernetwork to learn and adapt to the new tasks. 
  

The \textit{Fisher information matrix (FI)} provides the information on the  importance of each weight in the network. For the \textit{IWR} in \eqref{eq:4}, the FI matrix is defined as

\begin{small}
\begin{align}
   & FI^t =\frac{1}{N_t} \sum_{i=1}^{N_t} \biggl [ \nabla_{\Theta_m^t} L_{task}(\Theta_h, \mathbf{e}^t, y_i, \mathbf{x}_i) \times \nonumber \\ & \qquad \qquad \qquad \qquad \qquad \qquad  \nabla_{\Theta_m^t} L_{task}(\Theta_h, \mathbf{e}^t, y_i,  \mathbf{x}_i)^T \biggl ]
\end{align}
\end{small}
We note that the derivatives are computed with respect to the main network parameters to assess the importance of those parameters and not with respect to the hypernetwork parameters that we are learning unlike the standard regularization techniques. 
Thus, using Fisher information matrix we can find out the main network parameters which are important in learning the task. The existing regularisation for hypernetworks in \eqref{eq:1} treats all the main network parameters equally.  In practice, not all main network parameters contribute equally to solving a particular task. Hence, it is not required to exactly generate all the main network parameters by the hypernetwork but only important ones and can be achieved using the \textit{IWR} regularisation. The \textit{IWR} regularization is a generic technique that can be used with any hypernetwork and not only the \textit{LSTM} hypernetwork to improve the continual learning performance.


\subsection{Network Growth Technique}

One potential problem with the regularisation approaches is that training time grows with number of tasks as can be seen from Eq \eqref{eq:1} and Eq \eqref{eq:4}. Moreover, the same hypernetwork parameters are used to generate all the task specific main  parameters. This can become a bottleneck and affects continual learning performance in situations with a large number of tasks.  We propose \textit{LSTM} hypernetworks \textit{(LSTM\_NET\_GROW)}  based on network growth  for continual learning.  It provides more flexibility in adapting to the new task by maintaining task specific parameters and accelerates model training by requiring no regularisation. In order to transfer knowledge across the tasks, \textit{LSTM\_NET\_GROW} also maintains a shared set of hypernetwork parameters.

We hypothesize that,  though the actual  main network parameters differ across the tasks, the dependencies existing among the main network parameters remain the same across the tasks. Based on this intuition, we define the shared and task specific parameters in the \textit{LSTM} hypernetwork. In \textit{LSTM}, the dependencies are captured by the weights associated with the hidden state. Hence, we assume them to be the same across the tasks in the proposed \textit{LSTM\_NET\_GROW} model. The  variability in parameter generation across the tasks is captured by having task specific weights associated with inputs.   More specifically, the  weights (\(\mathbf{u_f,u_i,u_o,u_g}\)) of the \textit{LSTM} are shared across the tasks and we maintain input weights (\(\mathbf{w_f^t,w_i^t,w_o^t,w_g^t}\)) to be task specific.
\textit{LSTM\_NET\_GROW} uses the following \textit{LSTM} operations to generate chunk weights of the main network parameters associated with task $t$.

\begin{align*}
&\mathbf{i_j^{t}} =\sigma(\mathbf{w_{i}^t}\times{\langle\mathbf{e^t},\mathbf{c_j}\rangle}+\mathbf{u_i}\times \mathbf{h_{j-1}^{t}} ) \\
&\mathbf{f_j^{t}} =\sigma(\mathbf{w_{f}^t}\times{\langle\mathbf{e^t},\mathbf{c_j}\rangle}+\mathbf{u_f}\times\mathbf{h_{j-1}^{t} }) \\
&\mathbf{o_j^{t}} =\sigma(\mathbf{w_{o}^t}\times{\langle\mathbf{e^t},\mathbf{c_j}\rangle}+\mathbf{u_o}\times\mathbf{h_{j-1}^{t}} ) \\
&\mathbf{g_j^{t}} = Tanh(\mathbf{w_{g}^t}\times{\langle\mathbf{e^t},\mathbf{c_j}\rangle}+\mathbf{u_g}\times\mathbf{h_{j-1}^{t}} )\\
& \mathbf{s_j^{t}} = \mathbf{f_j^{t}} \odot \mathbf{s_{j-1}^{t}}  + \mathbf{i_j^{t}}  \odot \mathbf{g_j^{t}} \\
& \mathbf{h_j^{t}}  = \mathbf{o_j^{t}} \odot \tanh(\mathbf{s_j^t})\\
&\mathbf{\Theta_m^{(t,j)}}=\mathbf{h_j^{t} W^t} 
\end{align*}

\begin{small}

\begin{table*}
    \caption{Comparison of average test accuracy (\%) of Split MNIST and Permuted MNIST for all three scenarios of continual learning without generative replay}

  \begin{tabular}{ p{3.5cm}| p{1.8cm} p{1.8cm} p{2cm}| p{1.8cm} p{1.8cm} p{1.8cm}}
    \toprule
    &
      \multicolumn{3}{c|}{Split MNIST } &
      \multicolumn{3}{c}{Permuted MNIST }  \\
      \hline
      & {CL1} & {CL2} & {CL3} & {CL1} & {CL2} & {CL3} \\
      
      \midrule
    
    EWC \cite{10}&98.64\(\pm\)0.22 &63.95\(\pm\)1.90 &20.10\(\pm\)0.06& 94.74\(\pm\)0.05&94.31\(\pm\)0.11 &25.04\(\pm\)0.50\\
 
 {Online EWC}\cite{10}  & 99.12\(\pm\)0.11    &64.32\(\pm\)1.90& 19.96\(\pm\)0.07& 95.96\(\pm\)0.06 &94.42\(\pm\)0.13& 33.88\(\pm\)0.49\\
 
 SI\cite{11}&  99.09\(\pm\)0.15&65.36\(\pm\)1.57& 19.99\(\pm\)0.06& 94.75\(\pm\)0.14    &{95.33\(\pm\)0.11}&29.31\(\pm\)0.62\\
 
 {HNET}\cite{5}&99.79\(\pm\)0.01&87.01\(\pm\)0.47& 69.48\(\pm\)0.80&97.57\(\pm\)0.02&92.80\(\pm\)0.15& 91.75\(\pm\)0.21\\
 
 {HNET\_IWR}&99.79\(\pm\)0.01&88.51\(\pm\)0.18& 71.90\(\pm\)0.11 &97.60\(\pm\)0.04&93.90\(\pm\)0.11&92.15\(\pm\)0.19\\
 {LSTM\_NET}  &99.82\(\pm\)0.01 & 89.50\(\pm\)0.19&  71.31\(\pm\)0.07 &97.65\(\pm\)0.01 & 93.11\(\pm\)0.13  &  92.10\(\pm\)0.20\\
 
 {LSTM\_NET\_IWR}  &\textbf{99.85\(\pm\)0.02} & {90.17\(\pm\)0.25}&  {71.54\(\pm\)0.04}&{97.74\(\pm\)0.03} & 94.26\(\pm\)0.10&  {92.21\(\pm\)0.23}\\
  
 {LSTM\_NET\_GROW}  &\textbf{99.85\(\pm\)0.02} & \textbf{97.11\(\pm\)0.16}&  \textbf{83.21\(\pm\)0.02} &\textbf{97.88\(\pm\)0.02} & \textbf{95.46\(\pm\)0.09}&  \textbf{92.23\(\pm\)0.19}\\
    \bottomrule
  \end{tabular}
  \label{table:1}
\end{table*}
\end{small}

\begin{table*}
\caption{Comparison of average test accuracy (\%) of Split MNIST and Permuted MNIST for all three scenarios of continual learning with generative replay}
  \begin{tabular}{ p{3.5cm}| p{1.8cm} p{1.8cm} p{2cm}| p{1.8cm} p{1.8cm} p{1.8cm}}
    \toprule
     &
      \multicolumn{3}{c|}{Split MNIST } &
      \multicolumn{3}{c}{Permuted MNIST }  \\
      \hline
      & {CL1} & {CL2} & {CL3} & {CL1} & {CL2} & {CL3} \\
      \midrule
    LWF\cite{36}&99.57\(\pm\)0.02&71.50\(\pm\)1.63&23.85\(\pm\)0.44& 69.84\(\pm\)0.46&72.64\(\pm\)0.52 &22.64\(\pm\)0.23\\
DGR\cite{37}&99.50\(\pm\)0.03&95.72\(\pm\)0.25&90.79\(\pm\)0.41& 92.52\(\pm\)0.08& 95.09\(\pm\)0.04&92.19\(\pm\)0.09\\
DGR+distill\cite{37}&99.61\(\pm\)0.02&96.83\(\pm\)0.20&91.79\(\pm\)0.32&97.51\(\pm\)0.01&97.35\(\pm\)0.02&96.38\(\pm\)0.03\\
HNET+R\cite{5}& \textbf{99.83\(\pm\)0.01}&98.00\(\pm\)0.03&95.30\(\pm\)0.13& {97.87\(\pm\)0.01} &97.60\(\pm\)0.01&97.76\(\pm\)0.01\\
{HNET\_IWR+R}&\textbf{99.83\(\pm\)0.01}&97.94\(\pm\)0.05&95.38\(\pm\)0.16&97.85\(\pm\)0.01&97.66\(\pm\)0.02&97.76\(\pm\)0.02\\
LSTM\_NET+R  &\textbf{99.83\(\pm\)0.01} & 98.17\(\pm\)0.02&  95.46\(\pm\)0.11 & 97.87\(\pm\)0.01 & 97.60\(\pm\)0.01&  97.77\(\pm\)0.01\\
{LSTM\_NET\_IWR+R}  &\textbf{99.83\(\pm\)0.01} & {98.39\(\pm\)0.05}&  {96.50\(\pm\)0.19}&{97.87\(\pm\)0.01} & {97.66\(\pm\)0.01}& \textbf{97.80\(\pm\)0.02}\\
{LSTM\_NET\_GROW+R} &\textbf{99.83\(\pm\)0.01}&\textbf{98.43\(\pm\)0.05}&
\textbf{97.01\(\pm\)0.13} &\textbf{97.90\(\pm\)0.01}&\textbf{97.70\(\pm\)0.02}&\textbf{97.80\(\pm\)0.01}\\
    \bottomrule
  \end{tabular}
  
  \label{table:2}
\end{table*}

The \textit{LSTM\_NET\_GROW} model freezes the hidden weights (\(\mathbf{u_f,u_i,u_o,u_g}\)) of the \textit{LSTM}  after learning first task and is shared across all the tasks. It keeps learning new task specific input weights (\(\mathbf{w_f^t,w_i^t,w_o^t,w_g^t}\)) upon training on the new task and those are stored for inference at a later stage.  This approach does not require the additional regularisation term and can be learnt just based on task specific loss ($L_{task}$). In addition, the task specific weights provide extra flexibility to the \textit{LSTM} hypernetwork in generating task specific main network weights. 

\section{Experiments}
We perform extensive experiments on various continual learning setups and real world benchmarks datasets to show the effectiveness of our approach.
We present our results on Split MNIST, permuted MNIST, CIFAR-10, and CIFAR-100 datasets. Through experiments we aim to demonstrate: 
\begin{itemize} [noitemsep]
 \item impact of maintaining dependencies across the 
 chunks using the \textit{LSTM} based hypernetwork \textit{(LSTM\_NET)},
 
\item impact of the proposed regularisation \textit{IWR} on \textit{LSTM\_NET} (\textit{LSTM\_NET\_IWR}) and on HNET (\textit{HNET\_IWR}) in mitigating catastrophic forgetting,

\item improvement in performance using proposed dynamically growing \textit{LSTM} based hypernetwork \textit{LSTM\_NET\_GROW}. 
\item knowledge transfer and mitigating forgetting across the tasks using the challenging Cifar datasets. 
\end{itemize}

\subsection{Experimental Setup}
Continual learning models are tested on three different continual learning scenarios~\cite{35}. \\
\textit{CL1 (Task incremental learning)}: It provides the task identity information to the model both at training and testing time.  Since task identity is available, dedicated components can be assigned to the tasks in the sequence. Multi-headed model is one such kind of model architecture used in continual learning.  \\
\textit{CL2 (Domain incremental learning)}: It does not provide task identity information at test time, nor is it required to infer the task identity. Here, each task treats data to be from different domains but with the same classes. 
It considers the same number of classes across all the tasks in the sequence and utilises the same output heads for all the tasks. \\
\textit{CL3 (Class incremental learning)}: In this continual learning scenario, the model is not provided with the task identity.  It not only requires learning the tasks incrementally but also requires to infer the task identity. Here, the task identity is inferred through the predictive distribution entropy. This scenario resembles most with the real time setting with new class objects appearing incrementally.

 \begin{table*}[t]
    \caption{Test accuracy comparison of various methods on CIFAR-10 (C-10) and subsequent five splits ($S_{1\ldots5} $ ) each with ten classes of CIFAR-100 (C-100).}
\centering
    \begin{tabular}{p{3.5cm} p{1.05cm} p{1.05cm} p{1.05cm} p{1.05cm} p{1.05cm} p{1.05cm} p{2.3cm} } \toprule

 {}&{C-10 (\%)} & {C-100 $S_1$(\%)} & {C-100 $S_2$(\%)} &{C-100 $S_3$(\%)}&{C-100 $S_4$(\%)}&{C-100 $S_5$(\%)} & {Average-accuracy(\%)} \\ \midrule
{Finetuning}&15.3&13.1&12.2&10.2&20.3&87.0&26.35\\
{Training-from-scratch}&88.6& 79.3& 77.0&83.0&74.4&81.1&80.60\\
{HNET}\cite{5} & 88 & 83 & 79 & 82 & 81 & 82 &82.50 \\
{HNET\_IWR}&86.2&86.1&80.2&85.0&83.6&85.01&84.35\\
{LSTM\_NET\_during}&88.78&89.3&85.2&84.4&83.5&82.7&85.64\\
{LSTM\_NET}&88.74&89.1&84.9&84.3&83.4&82.7&85.52 \\
{LSTM\_NET\_IWR}&88.44&88.9&85.2&88.5&86.3&86.8&{87.35}\\
{LSTM\_NET\_GROW}&88.98&87.7&86.3&88.2&89.2&89.1&\textbf{88.25}\\
     \bottomrule
\end{tabular}
\label{table:3}
\end{table*}
The effectiveness of the hypernetwork based \textit{CL} techniques for parameter generation are  also tested on two continual learning setups, replay based and non-replay based setups. In the non-replay-based setup, hypernetwork is trained to generate parameters of the classifier used for solving the image classification problems. In the non-replay based setup, we compare the proposed approach with regularisation baselines \textit{Elastic Weight Consolidation (EWC)} \cite{10}, \textit{Synaptic Intelligence (SI)} \cite{11} and the baseline hypernetwork \textit{HNET}~\cite{5}. In the replay based setup, we augment our system with a generative model, for e.g. variational auto-encoder (\textit{VAE}) to generate synthetic examples from the previous tasks that can be replayed to aid the classifier to remember previous tasks. In this case, hypernetwork will generate weights for the replay network i.e. \textit{VAE} but not the target classifier. In the replay based setup, we compare the proposed approach with baselines \textit{deep generative replay with distillation (DGR)}\cite{37},  \textit{learning without forgetting (LWF)}\cite{36} and the baseline hypernetwork \textit{HNET}\cite{5}.

We conduct experiments on the standard continual learning task of image classification on publicly available real world data sets such as split MNIST, permuted MNIST, CIFAR-10 and CIFAR-100.  In these experiments, we use a single layer \textit{LSTM} that takes a task and chunk embedding each of size 96. For MNIST, we consider an \textit{LSTM} with hidden state size 64 and batch size 128. For CIFAR, hidden size and batch size are 128 and 32 respectively. For Split MNIST, the classifier is a fully connected network (FCN) with 2 layers each with size 400~\cite{5}. For Permuted MNIST, layer size is taken to be 1000 as done in \cite{5}. For CIFAR datasets, Resnet-32 is used as the classifier. In replay based setup, we use \textit{VAE} which uses an FCN with two layers each of size 400 both  as the encoder and  the decoder and uses a latent space of dimension 100.

\subsection{Results}
We demonstrate the results of our experiments on various image classification datasets used for the continual learning setup and several baselines. For fair comparison, we maintained the number of trainable parameters in the hypernetwork to be equal or lesser than the baseline methods.

\subsubsection{Split MNIST:}
    Split MNIST is a popular continual learning benchmark for image classification. The dataset consists of images of ten digits (0-9) and form five binary classification tasks by pairing them sequentially i.e.$\{(0,1),(2,3),(4,5),$ $(6,7), (8,9)\}$. The results are presented in Table~\ref{table:1} for non-replay based setup, and in Table~\ref{table:2} for the replay based setup. The results show the efficacy of our approach in achieving better continual learning performance in all the three CL scenarios and for each of the setups.
    
    The proposed hypernetwork \textit{LSTM\_NET} outperforms the baselines EWC, SI and HNET in the non-replay based setup, and outperforms the baselines LWF, DGR and HNET in the replay based setup.  The performance of \textit{HNET\_IWR} and \textit{LSTM\_NET\_IWR}  demonstrates that the proposed regularisation technique \textit{IWR} further improves the performance of \textit{HNET} and \textit{LSTM\_NET} in all the continual learning setups.   While the methods provide comparable results for the easier CL1 setting, the improvement in performance using the proposed techniques are more evident in the more complex and realistic CL2 and CL3 settings. This is even more evident in the non-replay setup where the standard techniques struggle. 
    We can observe that the proposed approach \textit{LSTM\_NET\_GROW} has significantly  improved the continual learning performance over the other models for these CL scenarios and setups. One of the major reasons for large improvement in accuracy of CL2 and CL3 with \textit{LSTM\_NET\_GROW} is because of dynamically expanding network with new tasks, this helps each task to have task-specific parameters which won’t get updated while learning new tasks and contribute significantly for improvement in performance.

\subsubsection{Permuted MNIST}

This CL benchmark is a variant of MNIST, it consists of tasks which are created by performing random permutation over MNIST images. The sequence of $T=10$ tasks are obtained by repeating this procedure. We consider a dataset with a sufficiently long sequence of tasks  ($T=10$) to investigate the remembering capacity of our continual learning model. The results presented in Table~\ref{table:1} and Table~\ref{table:2} for non-replay and replay based setups respectively demonstrates the effectiveness of the proposed approaches for continual learning on Permuted MNIST. The results follows a similar trend as in the split MNIST, with \textit{LSTM\_NET\_GROW}  giving the best results, followed by \textit{LSTM\_NET\_IWR} and \textit{LSTM\_NET}, beating the baseline approaches.

\subsubsection{CIFAR-10/100 dataset}

    We further evaluate the effectiveness of the proposed approaches on a more challenging image classification data CIFAR-10 and CIFAR-100.  The model is first trained on 10 classes of CIFAR-10, and subsequently on five sets of ten classes from cifar-100 following the experimental setup in \cite{5}. Thus, the model needs to learn $T=6$ tasks. We use ResNet-32 for classification of CIFAR-10/100 datasets and the hypernetworks are trained to generate parameters of ResNet-32 architecture.  The experiments are conducted on the CL1 scenario and non-replay based setup following \cite{5}. 
    In addition to the baseline hypernetwork~\textit{HNET}, we also consider baselines which allows us to demonstrate the knowledge transfer across the tasks. \textit{Training-from-scratch} baseline independently and separately learns the main network parameters for each task  and test the performance on the corresponding task. The baseline \textit{Finetuning}   adapts the main network to the new tasks  without taking into account catastrophic forgetting. The model adapted to the final task is then used for predicting performance on all the tasks.  
    To demonstrate effectiveness of \textit{LSTM\_NET} in dealing with catastrophic forgetting, we also considered a baseline \textit{LSTM\_NET\_during}, where we  test \textit{LSTM\_NET} on each task immediately after training on that task instead of testing after training on all the tasks as in \textit{LSTM\_NET}.
    
    We provide the results comparing all teh approaches in Table~\ref{table:3}.  We can clearly observe from the results that our approach \textit{LSTM\_NET} outperforms \textit{HNET} by a great margin on a challenging dataset like CIFAR-10/100.
    In Table \ref{table:3}, comparing results of \textit{LSTM\_NET},  \textit{LSTM\_NET\_IWR} and  \textit{LSTM\_NET\_GROW} with  \textit{Training-from-scratch},  we can see that the knowledge transfer across the tasks helps the proposed approaches in getting a better performance. We can also observe that the \textit{LSTM\_NET\_during} matches with \textit{LSTM\_NET} which indicates that \textit{LSTM} based hypernetwork is very effective in dealing with catastrophic forgetting.  We also perform experiments using different regularisation techniques. The proposed \textit{IWR} regularisation achieves a better result than the baseline regularisation proposed in \cite{5}. In fact, it improves the \textit{HNET} performance as well, an  improvement in the overall test accuracy by almost 2\%.  Hence, \textit{IWR} is an effective regularisation technique for any hypernetwork based continual learning approaches. The performance improves further by using the \textit{LSTM\_NET\_GROW} approach for continual learning in this data similar to MNIST.

\subsubsection{Ablation Study}

\begin{figure}[t]
    \centering
    \subfloat[Compression ratio effect\label{fig:3.1}]
    {\includegraphics[scale=0.43]{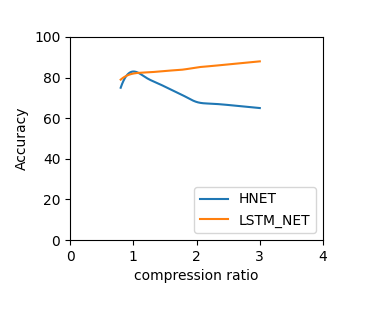}}\hfill
    \subfloat[Regularization effect\label{fig:3.2}]
    {\includegraphics[scale=0.43]{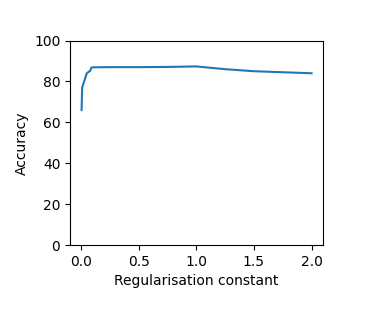}}
     \caption{In Figure \ref{fig:3.1}, plotted results of accuracies of \textit{LSTM\_NET} vs HNET with increase in compression ratio (ratio of trainable parameters in hypernetwork to main network) on Cifar datasets. In Figure \ref{fig:3.2},   
         plotted graph with accuracy values using \textit{LSTM\_NET\_IWR} with varying regularisation constant ($\beta$) in the IWR on Cifar datasets.}
    \label{fig:third}
\end{figure}

We conduct further ablation study on CIFAR-10/CIFAR-100 to understand the impact of compression ratio (Figure \ref{fig:3.1}) and regularisation constant (Figure \ref{fig:3.2}) on the proposed models. From Figure \ref{fig:3.1}, we can see that as the number of trainable parameters grows in hypernetwork compared to the main network, \textit{LSTM\_NET} performance further improves over the \textit{HNET}. In Figure \ref{fig:3.2}, we study the effect of varying the regularisation constant ($\beta$) in the \textit{IWR} regularization term in \textit{LSTM\_NET\_IWR}  on Cifar datasets. The performance is poor when regularization term is neglected (low value of $\beta$) as expected and is high and  stable for higher values of $\beta$.

From the experiments on CIFAR-10/CIFAR-100 tasks, we observe that training time using \textit{LSTM\_NET} with regularisations in Eq.\ref{eq:1} and Eq.\ref{eq:4} is approximately 28 hours but with \textit{LSTM\_NET\_GROW} it is approximately 20 hours. The above results are calculated with a batch size of 32 and 200 epochs for each batch. As the number of tasks increases, training time of regularisation approaches grows linearly but that of \textit{LSTM\_NET\_GROW} remains constant. On the other hand, the memory requirements of the regularisation approaches remains constant but \textit{LSTM\_NET\_GROW} grows linearly with tasks due to some task specific parameters but is much lower than maintaining separate parameters for each task.

\section{Conclusion}
We propose a novel \textit{LSTM} based hypernetwork for continual learning  which could capture  dependencies across the main network parameters while also maintaining the parameter efficiency. To improve continual learning performance using hypernetworks, we propose a novel regularisation \textit{Importance Weighted Regularisation (\textit{IWR})} which is well-suited for hypernetwork based CL approaches. To further improve the continual learning performance of the proposed \textit{LSTM}  hypernetwork, we propose a network growth technique for \textit{LSTMs}.  Through experiments on several image classification tasks and datasets, we demonstrate the effectiveness of our proposed approaches, $LSTM$ based hypernetwork, \textit{IWR} regularisation for hypernetworks, and network growth on \textit{LSTMs}. The proposed approaches improved the continual learning performance on all the CL tasks, settings, and datasets. As a future work, we would like to improve the parameter growth in \textit{LSTM\_NET\_GROW}, and develop hybrid models combining the network growth and regularisation to further improve CL performance.

{\small
\bibliographystyle{ieee_fullname}
\bibliography{egbib}

\begin{thebibliography}{10}\itemsep=-1pt

\bibitem{21}
Rahaf Aljundi, Francesca Babiloni, Mohamed Elhoseiny, Marcus Rohrbach, and
  Tinne Tuytelaars.
\newblock Memory aware synapses: Learning what (not) to forget.
\newblock In Vittorio Ferrari, Martial Hebert, Cristian Sminchisescu, and Yair
  Weiss, editors, {\em Computer Vision - {ECCV} 2018 - 15th European
  Conference, Munich, Germany, September 8-14, 2018, Proceedings, Part {III}},
  volume 11207 of {\em Lecture Notes in Computer Science}, pages 144--161.
  Springer, 2018.

\bibitem{24}
Arslan Chaudhry, Marc'Aurelio Ranzato, Marcus Rohrbach, and Mohamed Elhoseiny.
\newblock Efficient lifelong learning with {A-GEM}.
\newblock In {\em 7th International Conference on Learning Representations,
  {ICLR} 2019, New Orleans, LA, USA, May 6-9, 2019}. OpenReview.net, 2019.

\bibitem{26}
Arslan Chaudhry, Marcus Rohrbach, Mohamed Elhoseiny, Thalaiyasingam Ajanthan,
  Puneet~K Dokania, Philip~HS Torr, and Marc'Aurelio Ranzato.
\newblock On tiny episodic memories in continual learning.
\newblock {\em arXiv preprint arXiv:1902.10486}, 2019.

\bibitem{14}
Benjamin Ehret, Christian Henning, Maria~R. Cervera, Alexander Meulemans,
  Johannes von Oswald, and Benjamin~F. Grewe.
\newblock Continual learning in recurrent neural networks with hypernetworks.
\newblock {\em CoRR}, abs/2006.12109, 2020.

\bibitem{18}
Sibo Gai, Zhengyu Chen, and Donglin Wang.
\newblock Multi-modal meta continual learning.
\newblock In {\em International Joint Conference on Neural Networks, {IJCNN}
  2021, Shenzhen, China, July 18-22, 2021}, pages 1--8. {IEEE}, 2021.

\bibitem{27}
Chandan Gautam, Sethupathy Parameswaran, Ashish Mishra, and Suresh Sundaram.
\newblock Generalized continual zero-shot learning.
\newblock {\em CoRR}, abs/2011.08508, 2020.

\bibitem{6}
Stephen Grossberg.
\newblock Consciousness {CLEARS} the mind.
\newblock {\em Neural Networks}, 20(9):1040--1053, 2007.

\bibitem{4}
David Ha, Andrew~M. Dai, and Quoc~V. Le.
\newblock Hypernetworks.
\newblock In {\em 5th International Conference on Learning Representations,
  {ICLR} 2017, Toulon, France, April 24-26, 2017, Conference Track
  Proceedings}. OpenReview.net, 2017.

\bibitem{13}
Sepp Hochreiter and J{\"u}rgen Schmidhuber.
\newblock Long short-term memory.
\newblock {\em Neural computation}, 9(8):1735--1780, 1997.

\bibitem{34}
Ronald Kemker, Marc McClure, Angelina Abitino, Tyler~L. Hayes, and Christopher
  Kanan.
\newblock Measuring catastrophic forgetting in neural networks.
\newblock In Sheila~A. McIlraith and Kilian~Q. Weinberger, editors, {\em
  Proceedings of the Thirty-Second {AAAI} Conference on Artificial
  Intelligence, (AAAI-18), the 30th innovative Applications of Artificial
  Intelligence (IAAI-18), and the 8th {AAAI} Symposium on Educational Advances
  in Artificial Intelligence (EAAI-18), New Orleans, Louisiana, USA, February
  2-7, 2018}, pages 3390--3398. {AAAI} Press, 2018.

\bibitem{10}
James Kirkpatrick, Razvan Pascanu, Neil Rabinowitz, Joel Veness, Guillaume
  Desjardins, Andrei~A Rusu, Kieran Milan, John Quan, Tiago Ramalho, Agnieszka
  Grabska-Barwinska, et~al.
\newblock Overcoming catastrophic forgetting in neural networks.
\newblock {\em Proceedings of the national academy of sciences},
  114(13):3521--3526, 2017.

\bibitem{32}
Jan Koutn{\'{\i}}k, Faustino~J. Gomez, and J{\"{u}}rgen Schmidhuber.
\newblock Evolving neural networks in compressed weight space.
\newblock In Martin Pelikan and J{\"{u}}rgen Branke, editors, {\em Genetic and
  Evolutionary Computation Conference, {GECCO} 2010, Proceedings, Portland,
  Oregon, USA, July 7-11, 2010}, pages 619--626. {ACM}, 2010.

\bibitem{2}
Matthias~De Lange, Rahaf Aljundi, Marc Masana, Sarah Parisot, Xu Jia, Ales
  Leonardis, Gregory~G. Slabaugh, and Tinne Tuytelaars.
\newblock Continual learning: {A} comparative study on how to defy forgetting
  in classification tasks.
\newblock {\em CoRR}, abs/1909.08383, 2019.

\bibitem{36}
Zhizhong Li and Derek Hoiem.
\newblock Learning without forgetting.
\newblock In Bastian Leibe, Jiri Matas, Nicu Sebe, and Max Welling, editors,
  {\em Computer Vision - {ECCV} 2016 - 14th European Conference, Amsterdam, The
  Netherlands, October 11-14, 2016, Proceedings, Part {IV}}, volume 9908 of
  {\em Lecture Notes in Computer Science}, pages 614--629. Springer, 2016.

\bibitem{29}
Xialei Liu, Marc Masana, Luis Herranz, Joost van~de Weijer, Antonio~M.
  L{\'{o}}pez, and Andrew~D. Bagdanov.
\newblock Rotate your networks: Better weight consolidation and less
  catastrophic forgetting.
\newblock In {\em 24th International Conference on Pattern Recognition, {ICPR}
  2018, Beijing, China, August 20-24, 2018}, pages 2262--2268. {IEEE} Computer
  Society, 2018.

\bibitem{20}
Noel Loo, Siddharth Swaroop, and Richard~E. Turner.
\newblock Generalized variational continual learning.
\newblock In {\em 9th International Conference on Learning Representations,
  {ICLR} 2021, Virtual Event, Austria, May 3-7, 2021}. OpenReview.net, 2021.

\bibitem{23}
David Lopez{-}Paz and Marc'Aurelio Ranzato.
\newblock Gradient episodic memory for continual learning.
\newblock In Isabelle Guyon, Ulrike von Luxburg, Samy Bengio, Hanna~M. Wallach,
  Rob Fergus, S.~V.~N. Vishwanathan, and Roman Garnett, editors, {\em Advances
  in Neural Information Processing Systems 30: Annual Conference on Neural
  Information Processing Systems 2017, December 4-9, 2017, Long Beach, CA,
  {USA}}, pages 6467--6476, 2017.

\bibitem{38}
Michael McCloskey and Neal~J Cohen.
\newblock Catastrophic interference in connectionist networks: The sequential
  learning problem.
\newblock In {\em Psychology of learning and motivation}, volume~24, pages
  109--165. Elsevier, 1989.

\bibitem{7}
Martial Mermillod, Aur{\'e}lia Bugaiska, and Patrick Bonin.
\newblock The stability-plasticity dilemma: Investigating the continuum from
  catastrophic forgetting to age-limited learning effects, 2013.

\bibitem{1}
German~Ignacio Parisi, Ronald Kemker, Jose~L. Part, Christopher Kanan, and
  Stefan Wermter.
\newblock Continual lifelong learning with neural networks: {A} review.
\newblock {\em Neural Networks}, 113:54--71, 2019.

\bibitem{22}
Sylvestre{-}Alvise Rebuffi, Alexander Kolesnikov, Georg Sperl, and Christoph~H.
  Lampert.
\newblock icarl: Incremental classifier and representation learning.
\newblock In {\em 2017 {IEEE} Conference on Computer Vision and Pattern
  Recognition, {CVPR} 2017, Honolulu, HI, USA, July 21-26, 2017}, pages
  5533--5542. {IEEE} Computer Society, 2017.

\bibitem{25}
Matthew Riemer, Ignacio Cases, Robert Ajemian, Miao Liu, Irina Rish, Yuhai Tu,
  and Gerald Tesauro.
\newblock Learning to learn without forgetting by maximizing transfer and
  minimizing interference.
\newblock In {\em 7th International Conference on Learning Representations,
  {ICLR} 2019, New Orleans, LA, USA, May 6-9, 2019}. OpenReview.net, 2019.

\bibitem{30}
Andrei~A. Rusu, Neil~C. Rabinowitz, Guillaume Desjardins, Hubert Soyer, James
  Kirkpatrick, Koray Kavukcuoglu, Razvan Pascanu, and Raia Hadsell.
\newblock Progressive neural networks.
\newblock {\em CoRR}, abs/1606.04671, 2016.

\bibitem{31}
J{\"{u}}rgen Schmidhuber.
\newblock Learning to control fast-weight memories: An alternative to dynamic
  recurrent networks.
\newblock {\em Neural Comput.}, 4(1):131--139, 1992.

\bibitem{37}
Hanul Shin, Jung~Kwon Lee, Jaehong Kim, and Jiwon Kim.
\newblock Continual learning with deep generative replay.
\newblock In Isabelle Guyon, Ulrike von Luxburg, Samy Bengio, Hanna~M. Wallach,
  Rob Fergus, S.~V.~N. Vishwanathan, and Roman Garnett, editors, {\em Advances
  in Neural Information Processing Systems 30: Annual Conference on Neural
  Information Processing Systems 2017, December 4-9, 2017, Long Beach, CA,
  {USA}}, pages 2990--2999, 2017.

\bibitem{12}
Kenneth~O. Stanley, David~B. D'Ambrosio, and Jason Gauci.
\newblock A hypercube-based encoding for evolving large-scale neural networks.
\newblock {\em Artif. Life}, 15(2):185--212, 2009.

\bibitem{35}
Gido~M. van~de Ven and Andreas~S. Tolias.
\newblock Three scenarios for continual learning.
\newblock {\em CoRR}, abs/1904.07734, 2019.

\bibitem{5}
Johannes von Oswald, Christian Henning, Jo{\~{a}}o Sacramento, and Benjamin~F.
  Grewe.
\newblock Continual learning with hypernetworks.
\newblock In {\em 8th International Conference on Learning Representations,
  {ICLR} 2020, Addis Ababa, Ethiopia, April 26-30, 2020}. OpenReview.net, 2020.

\bibitem{28}
Jaehong Yoon, Eunho Yang, Jeongtae Lee, and Sung~Ju Hwang.
\newblock Lifelong learning with dynamically expandable networks.
\newblock In {\em 6th International Conference on Learning Representations,
  {ICLR} 2018, Vancouver, BC, Canada, April 30 - May 3, 2018, Conference Track
  Proceedings}. OpenReview.net, 2018.

\bibitem{11}
Friedemann Zenke, Ben Poole, and Surya Ganguli.
\newblock Continual learning through synaptic intelligence.
\newblock In Doina Precup and Yee~Whye Teh, editors, {\em Proceedings of the
  34th International Conference on Machine Learning, {ICML} 2017, Sydney, NSW,
  Australia, 6-11 August 2017}, volume~70 of {\em Proceedings of Machine
  Learning Research}, pages 3987--3995. {PMLR}, 2017.

\end{thebibliography}
}

\end{document}